# Statistical Characteristics of Driver Acceleration Behavior and Its Probability Model

Rui Liu[1], Xuan Zhao[1], Xichan Zhu[2*], Jian Ma[1]
1. School of Automobile, Chang'an University, Xi'an 710064, China
2. School of Automotive Studies, Tongji University, Shanghai 201804, China

**Abstract:** Naturalistic driving data were applied to study driver acceleration behaviour, and a probability model of the driver was proposed. First, the question of whether the database is large enough is resolved using kernel density estimation and Kullback-Liebler divergence. Next, the convergence database is utilised to achieve the bivariate acceleration distribution pattern. Subsequently, two probability models are proposed to explain the pattern. Finally, the statistical characteristics of the acceleration behaviours are studied to verify the probability models. The longitudinal and lateral acceleration behaviours always approximate a similar Pareto distribution. The braking, accelerating, and steering manoeuvres become more intense at first and then less intense as the velocity increases. These behaviours characteristics reveal the mechanism of the quadrangle bivariate acceleration distribution pattern. The bivariate acceleration behaviour of the driver will never reach a circle-shaped pattern. The bivariate Pareto distribution model can be applied to describe the bivariate acceleration behaviour of the driver.

**1. Introduction**

With the increase in road testing and commercial applications of driving automation systems, there have been many fatal accidents related to intelligent vehicles in recent years. These accidents indicate that the driver should continue monitoring the surrounding environment and maintain control of the vehicle until autonomous driving technology is sufficiently safe [1]. When the driver and driving automation system jointly control the vehicle, the system needs to consider the driver behaviour. First, the driving automation system should keep the driver comfortable. The divergence between the driver and the system will not only reduce the driver's trust in an intelligent vehicle [2], but also bring many other unpredictable safety problems [3]. Second, the driving automation system should consider the interaction with the surrounding traffic in a mixed traffic flow. The braking operation of a driver is much slower than that of a driving automation system [4]. In the car-following scenario, a rear-end collision risk occurs if the leading intelligent vehicle brakes too fast. All these require a driving automation system to have the capability of human-like driving. To improve the human-like driving ability of intelligent vehicles, it is essential to study driver behaviour. The driver's behaviour at a certain moment is affected by the surrounding traffic environment, mental state of the driver, etc. However, the behaviour characteristics exhibit some regular features if the driver behaviour is placed within a longer time scale. Hence, the statistical characteristics of the driver behaviour were studied.

Acceleration can be applied to evaluate a driver's behaviour [5]. Human drivers control the vehicle mainly by three manoeuvres: steering the steer wheel, stepping on the accelerator pedal, and stepping on the brake pedal. These behaviours are directly correlated with the acceleration of the vehicle. The longitudinal acceleration is directly related to the manoeuvres of stepping on the accelerator or brake pedal, whereas the lateral acceleration is directly related to the steering manoeuvre. Therefore, studies on accelerations can help increase the human-like driving ability of the driving automation system.

*Corresponding Author: E-mail: zhuxichan@tongji.edu.cn. Address: No. 4800, Cao An Highway, Shanghai 201804, China.

The probability acceleration behaviour model of the driver has an important position in traffic simulation, to realize vehicle state estimation. The exponential distribution model was employed to describe the prior distribution of a driver's acceleration behaviour, and this probability model was utilised in a Monte Carlo simulation to assess the risk level [6, 7]. However, these studies did not explain why the exponential distribution was chosen as the prior model of the acceleration behaviour of the driver. Moreover, the acceleration behaviour characteristics of the driver are crucial for vehicle control. The tire friction circle is the physical limit of the acceleration [8]. Additionally, the friction circle [9, 10] or the oval subset of the friction circle [11, 12] was considered as the boundary of the driver acceleration behaviour in some early studies. Nevertheless, recent studies have shown that driver acceleration behaviour obey unique laws rather than friction limits [13, 14]. Accelerations may never reach the physical limit in daily driving. Furthermore, the acceleration behaviour can be applied to study the driving mechanism of the driver. The longitudinal and lateral driver acceleration behaviours are mutually influential, and the acceleration behaviours are influenced by the driving speed. It was shown that the peak lateral acceleration decreased as the longitudinal velocity increased when the vehicle was driving on a curved road [15, 16]. Hence, the statistical characteristics of driver acceleration behaviours are studied, and a probability model of the driver is proposed based on the acceleration behaviours.

This paper is organized as follows: phenomenon display, model hypothesis, behaviour study, and model verification. Section 2 introduces the naturalistic driving data and determines whether the database is large enough. Next, the bivariate acceleration distribution pattern of the driver is presented in section 3.1. Two probability models are proposed in section 3.2. The acceleration behaviour characteristics of the driver are studied, and the probability models are verified according to the acceleration behaviours in sections 3.3 through to section 3.5. Finally, the results are discussed in section 4, followed by the conclusion in section 5.

**2. How much data are enough?**

2.1 Naturalistic driving studies

The naturalistic driving data from the China-FOT (China Field Operational Test) were used in this study. Thirty-two drivers participated in the test. There were 25 men and seven women. The average age of the drivers was 32 years (SD: 2.84, range: 28–39). The drivers all had their own vehicles and had more than three years of driving experience. The annual driving mileages of the drivers averaged 17,875 km (*SD*: 6,288.96, range 5,000–30,000). The test vehicles were Volvo S60 equipped with various sensors, and were given to the drivers for three months. No limits on when or where the vehicles may be used were imposed. Most of the data were collected on urban, rural, urban-elevated roads, and freeways in Shanghai. The data sampling frequency in the China-FOT was 10 Hz. All the available driving data in the China-FOT were applied to constitute the database in this study, which is denoted as *Ω*. The quantity of observational data in *Ω* was 123,558,489 datapoints. The travel distance was 121,951 km, and the travel time was 3,432h.

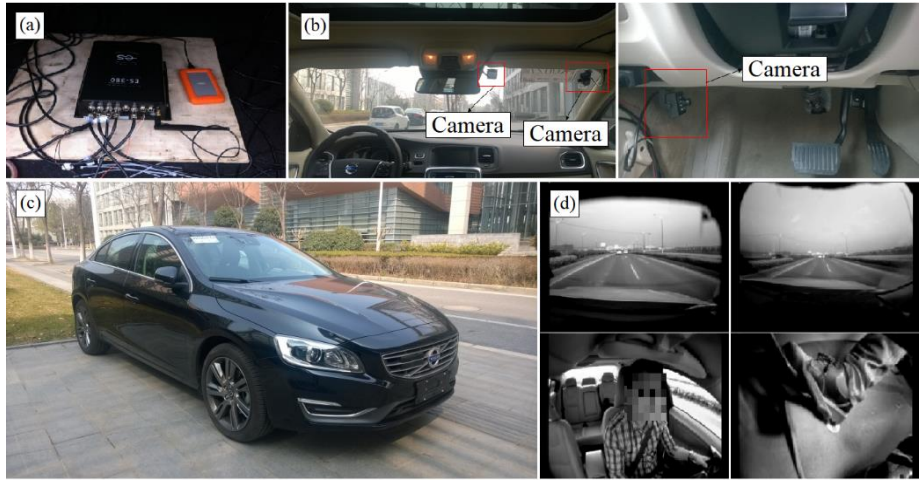

Figure 1. China-FOT: (a) data acquisition system, (b) camera installation location, (c) test vehicle, and (d) video information.

2.2. The convergences of the acceleration behaviours

A primary issue before the naturalistic driving data (NDD) is applied to study the driver acceleration behaviour is how to determine whether the database is large enough. Only by ensuring that the NDD is sufficient to obtain a convergent driver acceleration behaviour can a convincing conclusion be reached. Moreover, the quantity of the NDD should not be too high because the collection of NDD is extremely time-consuming and costly. Research on 'How much NDD can obtain credible driving behaviour of the driver?' are very limited. A statistical approach [17] was improved and was employed to estimate the appropriate quantity of NDD.

2.2.1 Kernel density estimation

The Gaussian mixture model (GMM) [18] and kernel density estimation (KDE) [19] are two commonly used methods that utilise a limited set of observational data to estimate empirical distribution. The distributions of the accelerations were estimated using the KDE. The KDE of a series of $d$-dimensional observational vectors ($x_i$) can be defined as

$$\hat{f}_n(\boldsymbol{x}) = \frac{1}{n}\sum_{i=1}^{n} \boldsymbol{K}(\boldsymbol{x}-\boldsymbol{x}_i) \tag{1}$$

Where $n$ is the quantity of $x_i$, and $K(x)$ is the kernel function.

The Gaussian kernel function is chosen as

$$\boldsymbol{K}(\boldsymbol{x}) = (2\pi)^{-d/2}\left|\boldsymbol{H}\right|^{-1/2}\exp(-\frac{1}{2}\boldsymbol{x}^{\mathrm{T}}\boldsymbol{H}^{-1}\boldsymbol{x}) \tag{2}$$

where $H$ is the matrix of the bandwidth and $|H|$ is the determinant of the matrix.

The choice of the bandwidth matrix has a significant influence on the KDE precision. The bandwidth matrix in KDE cannot be solved directly. Therefore, three methods were identified to solve the bandwidth: rule of thumb selector, plug-in selectors, and cross-validation selectors. The rule of thumb bandwidth selector of Silverman [20] works well for a dataset that follows a normal distribution. However, this method cannot be used when the dataset does not follow a normal distribution. Data-based automatic bandwidth selection methods, including plug-in selectors [21, 22] and cross-validation selectors [23-25], can be applied to estimate the bandwidth of a dataset that does not follow a normal distribution. Hence, the plug-in selector [22] is chosen to estimate the bandwidth.

2.2.2 Kullback-Leibler divergence

The similarity between the probability distributions of the two datasets can be measured using the Kullback-Leibler (KL) divergence [26]. Assuming that there are $n$ observational data in one dataset and $n+m$ observational data in another dataset, the KL divergence is defined as

$$D_{KL}[\hat{f}_{n+m}(\boldsymbol{x}) \| \hat{f}_n(\boldsymbol{x})] = \int_{-\infty}^{\infty} \cdots \int_{-\infty}^{\infty} \hat{f}_{n+m}(\boldsymbol{x}) \log \frac{\hat{f}_{n+m}(\boldsymbol{x})}{\hat{f}_n(\boldsymbol{x})} d\boldsymbol{x} \tag{3}$$

The $D_{KL}$ indicates the change of the distribution of one dataset after a new set of data is added to it. When the distribution tends to converge, the $D_{KL}$ will be sufficiently small in the process of continuously adding new data to the former dataset. If there is an $\Gamma$ that satisfies equation (4), $\Gamma$ is defined as the quantity of data that can obtain a convergent distribution.

$$\forall \Gamma \leq n \leq N, \left\| D_{KL}[\hat{f}_{n+m}(\boldsymbol{x}) \| \hat{f}_n(\boldsymbol{x})] \right\| < \varepsilon \tag{4}$$

Where $N$ is the maximum amount of data in $\Omega$. $\varepsilon$ is the threshold.

2.2.3 Examination process and results

The algorithm to check whether the database is large enough to acquire the convergent acceleration behaviour based on the KDE and $D_{KL}$ is shown below.

Convergence examination algorithm:

1) $1 \times 10^5$ observational datapoints are randomly chosen to compose an initial dataset.

2) A $1 \times 10^5$ set of new observational data are added to the former dataset.

3) The KDE of the former dataset and the KDE of the latter dataset are calculated. The $D_{KL}$ of these two datasets can be achieved.

4) If equation (4) is not satisfied, set $k=k+1$ and go to step 2); if equation (4) is satisfied and $N-k \geq 1 \times 10^7$, success and stop, and set $\Gamma=k$; if equation (4) is satisfied and $N-k < 1 \times 10^7$, fail and stop.

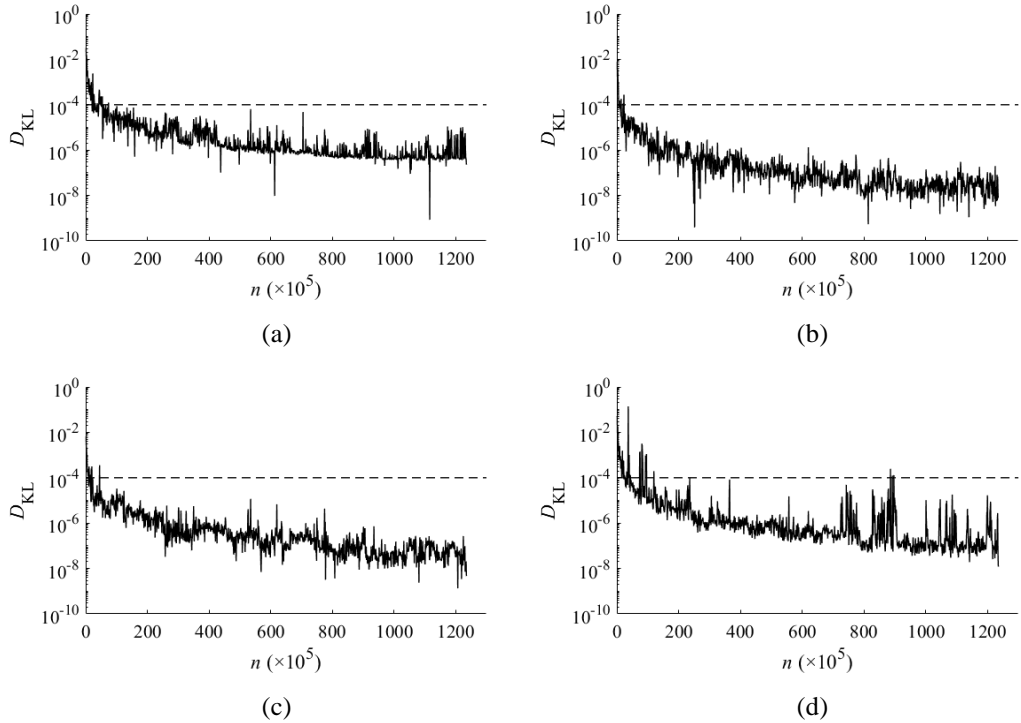

Figure 2. Kullback-Leibler divergence: (a) bivariate accelerations, (b) longitudinal acceleration, (c) lateral acceleration and (d) longitudinal velocity.

The value of $\varepsilon$ has a direct impact on the results. To determine if the acceleration distribution truly converged, a conservative value of $\varepsilon$ is selected, that is, $\varepsilon=10^{-4}$. This choice is consistent with that in the previous work [17]. Furthermore, the condition in equation (4) indicates that the $D_{KL}$ of two adjacent datasets should never be larger than the threshold in the following process. If this process is too short, it will be difficult to judge whether the distribution is truly converging. The condition that the $D_{KL}$ remains smaller than the threshold in more than 100 steps is used to ensure that the database is sufficient, that is, $N-k \geq 1 \times 10^7$. This condition means that the distribution has no significant change after $1 \times 10^7$ new observational data are added.

Table 1 Quantity of data required to obtain convergent behaviours

|  | observational vector ($\boldsymbol{x}$) | $D_{KL}$ | $\Gamma$ |
|---|---|---|---|
| bivariate accelerations | $[a_x, a_y]^T$ | Figure 2(a) | $0.74 \times 10^7$ |
| longitudinal acceleration | $a_x$ | Figure 2(b) | $0.23 \times 10^7$ |
| lateral acceleration | $a_y$ | Figure 2(c) | $0.45 \times 10^7$ |
| longitudinal velocity | $v_x$ | Figure 2(d) | $8.97 \times 10^7$ |

The convergence examination algorithm was utilised to determine the appropriate quantity of data required to obtain convergent behaviours for: the bivariate joint distribution of accelerations, univariate distributions of longitudinal and lateral accelerations, and univariate distribution of longitudinal velocity. The quantities of data required for convergent acceleration distributions are always less than $1 \times 10^7$, whereas the quantity of data required to obtain a convergent velocity distribution is less than $9 \times 10^7$. The quantity of data needed to obtain a convergent velocity distribution is much larger than those of the accelerations. The reason for this will be demonstrated in the upcoming sections, suffice to say that it is due to the velocity following an entirely different univariate distribution. These results demonstrate that the database $\Omega$ can acquire convergent distributions of the accelerations and velocity.

**3. Acceleration behaviour and probability model**

In this section, the database $\Omega$ was applied to analyse the bivariate acceleration behaviour. First, a bivariate acceleration distribution pattern was presented. Next, two bivariate distribution models were proposed to explain the pattern. Finally, the statistical characteristics of the acceleration behaviours were studied to illustrate the mechanism of the bivariate acceleration distribution pattern, and the bivariate distribution models were verified according to the acceleration behaviours.

3.1 Bivariate acceleration distribution pattern

The bivariate acceleration distribution pattern, with the relative density contours is shown in Figure 3. The boundary of the acceleration behaviour is chosen as the 0.0001% relative density contour, which indicates that only 10 parts per million (ppm) of the driving data (Table 2) remain in the outer area of this boundary.

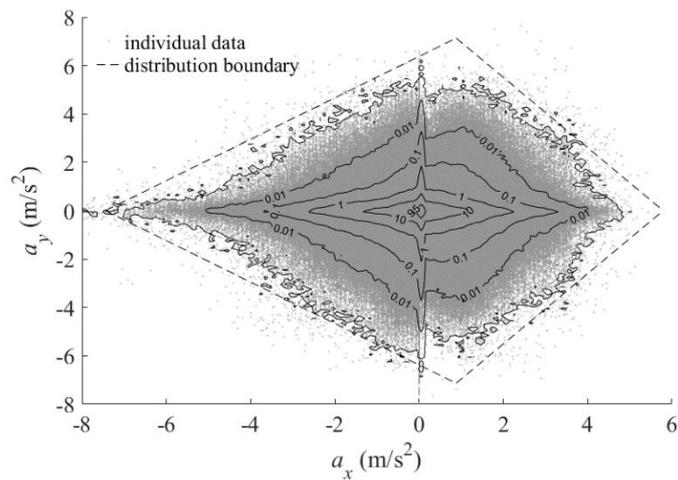

Figure 3. Bivariate acceleration distribution pattern.

Table 2 Density contour and data percentile

| Relative density contour | Percentage of data in the inner area of the contour | Approximate data percentile |
|---|---|---|
| 0.0001% | 99.997% | 99.999th |
| 0.01% | 99.854% | 99.9th |
| 0.1% | 98.960% | 99th |
| 1% | 94.639% | 95th |
| 10% | 76.450% | 75th |
| 95% | 26.949% | 25th |

The bivariate acceleration distribution presents a quadrangle pattern, that is, four closed lines can be used to fit the boundary of the bivariate acceleration behaviour. The distribution area of the brake deceleration is significantly greater than that of the forward acceleration, whereas the lateral accelerations on the left and right are approximately symmetric. In the following sections, the brake deceleration ($a_{x,b}$) and forward acceleration ($a_{x,f}$) are separated, whereas the left and right lateral accelerations are not distinguished.

3.2 Two bivariate distribution model hypotheses

Two representative bivariate distribution models were proposed to explain the bivariate acceleration distribution pattern, that is, the bivariate normal distribution model (BNDM) and the bivariate Pareto distribution model (BPDM).

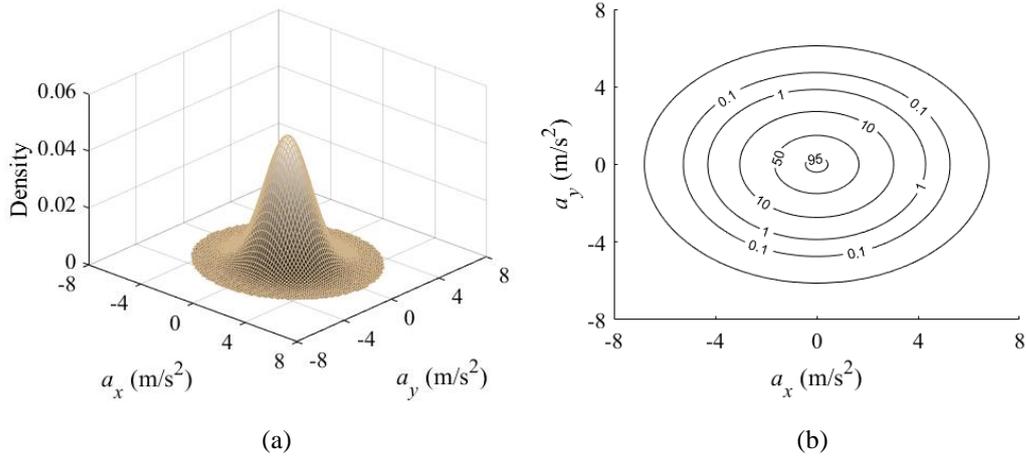

Figure 4. Bivariate normal distribution model: (a) bivariate distribution and (b) relative density contour.

First, it is assumed that the bivariate acceleration behaviour follows a bivariate normal distribution, which can be denoted as $N(\mu_x, \mu_y, \sigma_{N,x}^2, \sigma_{N,y}^2, \rho)$. It is set such that $\mu_x=0$, $\mu_y=0$, and $\rho=0$. These are reasonable assumptions because the centre point of the longitudinal acceleration or lateral acceleration is always at zero and will not move. Therefore, the probability density function (PDF) of the BNDM is

$$f_N(x, y) = \frac{1}{2\pi\sigma_{N,x}\sigma_{N,y}}\exp(-\frac{x^2}{\sigma_{N,x}^2}-\frac{y^2}{\sigma_{N,y}^2}) \tag{5}$$

where $x$ denotes the forward acceleration or the absolute value of the brake deceleration. $y$ denotes the lateral acceleration.

At some PDF plane where $f_N(x,y)=C$, the contour equation of the BNDM is

$$\frac{x^2}{\sigma_{N,x}^2}+\frac{y^2}{\sigma_{N,y}^2}=\eta \tag{6}$$

Where, $\eta=-\ln(2\pi\sigma_{N,x}\sigma_{N,y}C)$, with $C$ as a constant.

The contours of the BNDM are elliptical (Figure 4(b)). The marginal and conditional distributions of the BNDM follow the same univariate normal distribution.

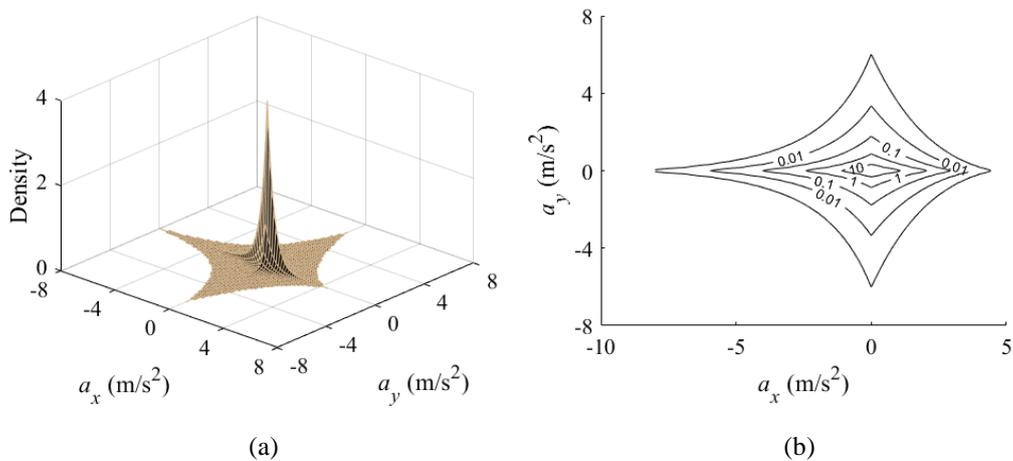

Figure 5. Bivariate Pareto distribution model: (a) bivariate distribution and (b) relative density contour.

Next, the BPDM is evaluated. It is assumed that the longitudinal acceleration behaviour always follows a Pareto distribution whose distribution parameters are the same at different lateral acceleration intervals, and vice versa. Hence, the conditional distributions are

$$f_{P,x}(x \mid y_1 \leq y \leq y_2) = \frac{1}{\sigma_{P,x}}(1+k_x\frac{x}{\sigma_{P,x}})^{-1-\frac{1}{k_x}}$$
$$f_{P,y}(y \mid x_1 \leq x \leq x_2) = \frac{1}{\sigma_{P,y}}(1+k_y\frac{y}{\sigma_{P,y}})^{-1-\frac{1}{k_y}}$$
(7)

The two variables are independent of each other. Therefore, the marginal distributions of the variables are

$$f_{P,x}(x) = f_{P,x}(x \mid y_1 < y < y_2)$$
$$f_{P,y}(y) = f_{P,y}(y \mid x_1 < x < x_2)$$
(8)

The PDF of the BPDM is

$$f_P(x,y) = f_{P,x}(x) \cdot f_{P,y}(y) \tag{9}$$

At some PDF plane where $f_P(x,y)=C$, the contour equation of the BPDM can be described as

$$y = \frac{1}{\lambda_y}[\omega_y(1+\lambda_x x)^\gamma - 1] \tag{10}$$

Where, $\lambda_x = \frac{k_x}{\sigma_{P,x}}$, $\lambda_y = \frac{k_y}{\sigma_{P,y}}$, $\omega_y = (\sigma_{P,x}\sigma_{P,y}C)^{-\frac{k_y}{1+k_y}}$, and $\gamma = -\frac{k_y(k_x+1)}{k_x(k_y+1)}$.

The contour equation of the BPDM is a polynomial function. The marginal and conditional distributions of the BPDM follow the same univariate Pareto distribution. By carefully choosing the distribution parameters (Table 4), the contours of the BPDM are very similar to those of the bivariate acceleration distribution obtained using the NDD (Figure 5(b)).

3.3 Univariate acceleration behaviour and model verification

It is difficult for existing methods to quantitatively directly study bivariate distribution features. Therefore, the marginal and conditional distributions of the bivariate acceleration behaviour were studied. If the marginal and conditional distributions of the acceleration behaviour are the same as those in the bivariate distribution model, then it can be considered that the bivariate distribution model can be employed to describe the driver acceleration behaviour.

The database $\Omega$ was used to analyse the univariate acceleration behaviour characteristics. The empirical distribution was acquired using the KDE method. The MATLAB Statistics and Machine Learning Toolbox [27] was utilised to study the empirical distribution, which evaluates the fitting results of 17 different kinds of statistical distribution models. The Akaike information criterion (AIC) and Bayesian information criterion (BIC) [28, 29] were employed to assess the goodness-of-fit. The AIC is defined as

$$\text{AIC} = 2r - 2\ln(L)$$
$$L = \hat{f}(x \mid \theta, M)$$
(11)

where $r$ is the number of statistical parameters, $x$ is the observational data, $L$ is the maximised value of the likelihood function of the statistical distribution $M$, and $\theta$ is the statistical parameter of $M$.

The BIC is defined as

$$\text{BIC} = \ln(n)r - 2\ln(L)$$
$$L = \hat{f}(x \mid \theta, M)$$
(12)

Where *n* is the quantity of the observational data.

The fitting results show that the Pareto distribution model is always optimal for braking deceleration, forward acceleration, and lateral acceleration. The fitting results of three kinds of statistical models and the empirical distributions are shown in Figure 6. The normal distribution model performs poorly in fitting the empirical distribution of the lateral acceleration. First, the probability density of the normal distribution model is too small in the place where the lateral acceleration is close to zero. Second, the probability density of the normal distribution model drops rapidly when the lateral acceleration increases. Table 3 lists the AIC and BIC values of these three statistical models. The AIC and BIC of the Pareto distribution model and the exponential distribution model are much smaller than those of the normal distribution model. These results are also correct for brake deceleration behaviour and forward acceleration behaviour. Therefore, the Pareto distribution model is the most appropriate model to describe the univariate driver acceleration behaviour, whereas the acceleration behaviour cannot be described using the normal distribution model. This reveals that the marginal distributions of the accelerations approximate the Pareto distribution. The fitting parameters of the Pareto distribution are presented in Table 4.

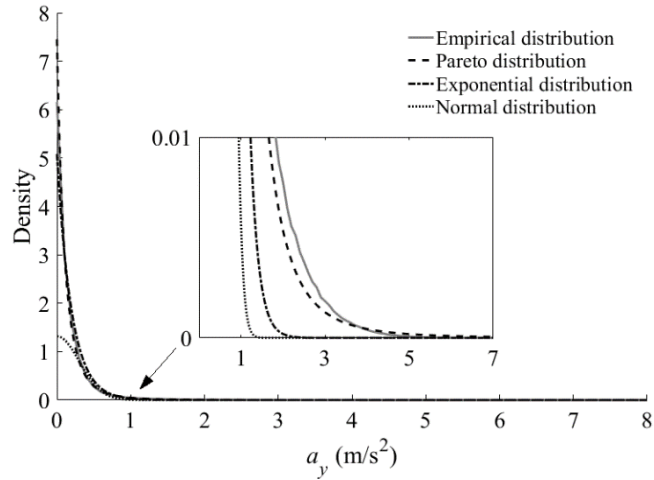

Figure 6. Three statistical models and empirical distribution.

Table 3 Fitting results of different statistical distributions

| Statistical Distribution | PDF | AIC | BIC |
| --- | --- | --- | --- |
| Normal Distribution | $f(x)=\dfrac{1}{\sqrt{2\pi}\sigma}e^{-\frac{(x-\mu)^2}{2\sigma^2}}$ | $5.54\times10^7$ | $5.54\times10^7$ |
| Pareto Distribution | $f(x)=\dfrac{1}{\sigma}(1+k\dfrac{x}{\sigma})^{-1-\frac{1}{k}}$ | $-1.71\times10^8$ | $-1.71\times10^8$ |
| Exponential Distribution | $f(x)=\dfrac{1}{\mu}e^{-\frac{x}{\mu}}$ | $-1.55\times10^8$ | $-1.55\times10^8$ |

Table 4 Pareto distribution fitting parameters

| Section | Fitting Parameters | Pareto distribution parameters in Figure 5* |
| --- | --- | --- |
| Left lateral acceleration | $k_y$=0.2978, $\sigma_{P,y}$=0.1370 | $k_y$=0.3, $\sigma_{P,y}$=0.136 |
| Right lateral acceleration | $k_y$=0.3177, $\sigma_{P,y}$=0.1356 | |
| Forward acceleration | $k_x$=-0.0429, $\sigma_{P,x}$=0.5063 | $k_x$=-0.043, $\sigma_{P,x}$=0.47 |
| Brake deceleration | $k_x$=0.0894, $\sigma_{P,x}$=0.4544 | $k_x$=0.09, $\sigma_{P,x}$=0.47 |

* The percentages of data in the four quadrants were assumed to be uniform.

The fitting results also show that the univariate distributions of the lateral acceleration at different longitudinal acceleration intervals always approximate a similar Pareto distribution, and vice versa. Figure 7 shows the examples where the univariate lateral acceleration behaviours approximate similar Pareto distributions at different longitudinal acceleration intervals. This reveals that the conditional distributions of the accelerations at different intervals always approximate the Pareto distribution.

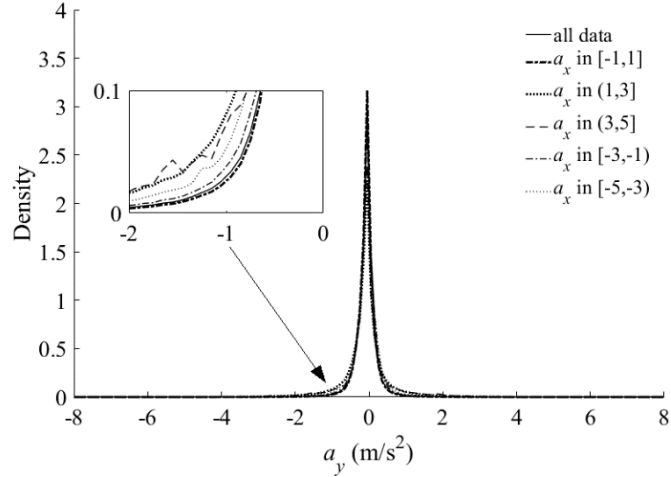

Figure 7. Univariate distributions of lateral acceleration in different longitudinal acceleration intervals.

The marginal distributions of the accelerations approximate the Pareto distribution (Figure 6). In addition, the conditional distributions of the accelerations at different intervals always approximate a similar Pareto distribution (Figure 7). Moreover, Figure 5(b) shows that the density contours of the BPDM are very similar to those obtained using the NDD. Hence, the bivariate acceleration behaviour can be well-described using the BPDM.

3.4 Interactive characteristics of the acceleration behaviours

Figure 3 and Figure 5(b) have apparent differences. This is due to the assumption in the BPDM that the longitudinal and lateral acceleration behaviours are independent of each other. Figure 7 indicates that the univariate lateral acceleration behaviours always exhibit some differences in different longitudinal acceleration intervals. This shows that the longitudinal and lateral acceleration behaviours are not mutually independent. These differences significantly influence the contours of the bivariate Pareto distribution. Hence, the percentiles of the accelerations at different intervals were analysed to identify the interactions between the longitudinal and lateral acceleration behaviours.

Generally, more than 50% of the data are distributed in the interval where the acceleration is close to zero. For example, 61% of the observational data are distributed in the interval where the lateral acceleration ranges from 0 to 0.2 m/s$^2$. Meanwhile, the Pareto distribution indicates that the data in the large lateral acceleration region cannot be ignored. Therefore, the 90th to 99.99th percentiles of the accelerations were presented to analyse the trend of the acceleration behaviour. The percentiles of the lateral acceleration at different longitudinal acceleration intervals are shown in Figure 8(a). The percentiles of brake deceleration and forward acceleration at different lateral acceleration intervals are shown in Figure 8(b) and 8(c). It is identified that the percentiles always move up as the other acceleration increases, that is, the Pareto distribution becomes more dispersive. The interactive characteristics between the longitudinal and lateral acceleration behaviours explain the differences between the BPDM and NDD. Hence, the reasons for the quadrangle pattern of the bivariate acceleration behaviour are as follows: (1) the bivariate Pareto distribution and (2) the interactions between the longitudinal and lateral acceleration behaviours.

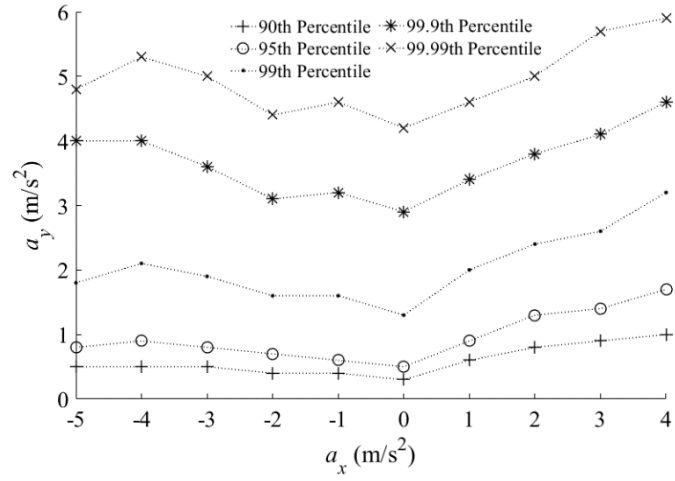

(a)

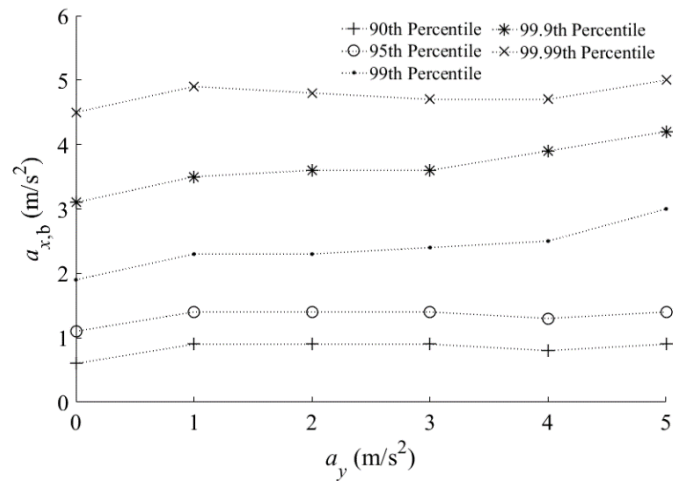

(b)

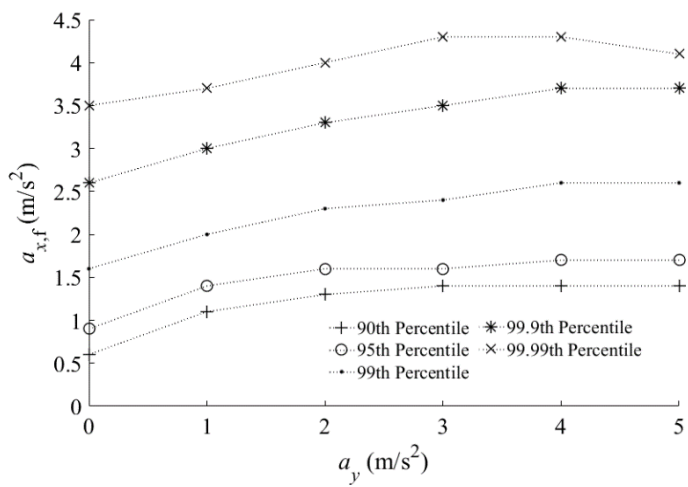

(c)

Figure 8. Percentiles in different acceleration intervals: (a) lateral acceleration in different longitudinal acceleration intervals, (b) brake deceleration in different lateral acceleration intervals and (c) forward acceleration in different lateral acceleration intervals.

3.5 Influence of the velocity on acceleration behaviours

Velocity is a very important vehicle state parameter. Drivers will choose the proper velocity based on the current driving scenarios, and will also adjust their driving behaviour according to the velocity. Therefore, the database $\Omega$ was applied to analyse the acceleration behaviour of the driver at different velocity intervals. The empirical distribution of the velocity is shown in Figure 9. The PDF is very large when the velocity approaches 0 m/s. The PDF is approximately horizontal when the velocity ranges from 0 to 15 m/s, and the PDF linearly decreases to 0 when the velocity is larger than 15 m/s.

Univariate distributions of the accelerations at different velocity intervals were analysed. It is found that the lateral acceleration behaviour, brake deceleration behaviour, and forward acceleration behaviour always follow the Pareto distribution in each velocity interval. Similarly, the 90th to 99.99th percentiles of the accelerations were studied. The percentiles and relative density contours of the lateral acceleration, brake deceleration, and forward acceleration are shown in Figure 10. The relative density contours in these figures are achieved using the same method as in Figure 3. It is shown that the percentiles are highly consistent with the relative density contours. The changes in the percentiles are mainly due to the different parameters of the Pareto distribution at different velocity intervals. This indicates that the driver acceleration behaviour is intense at medium speed (5 m/s–10 m/s), whereas the driver acceleration behaviour is relatively gentle at high speed or low speed.

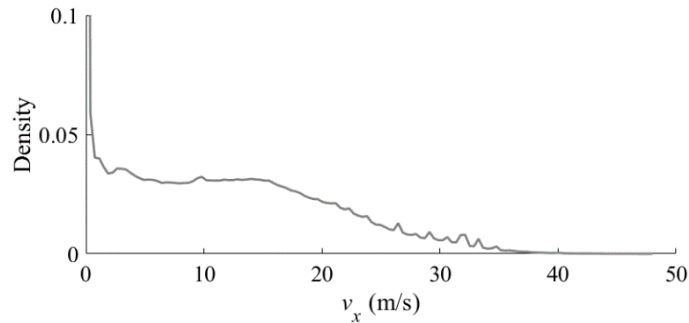

Figure 9. Empirical distribution of the velocity.

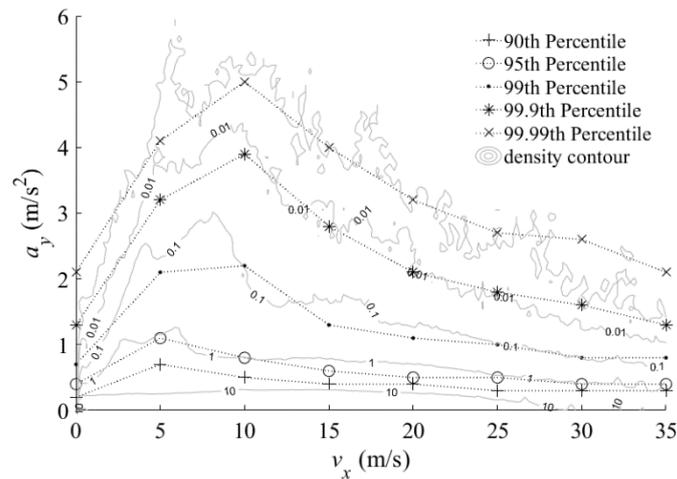

(a)

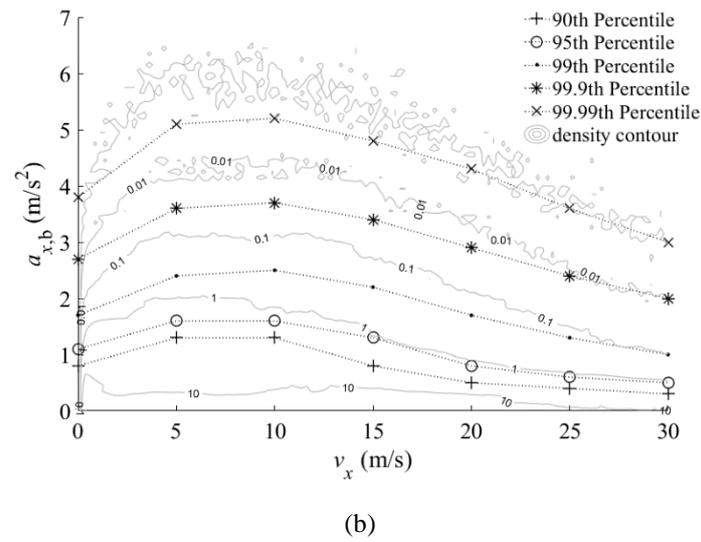

(b)

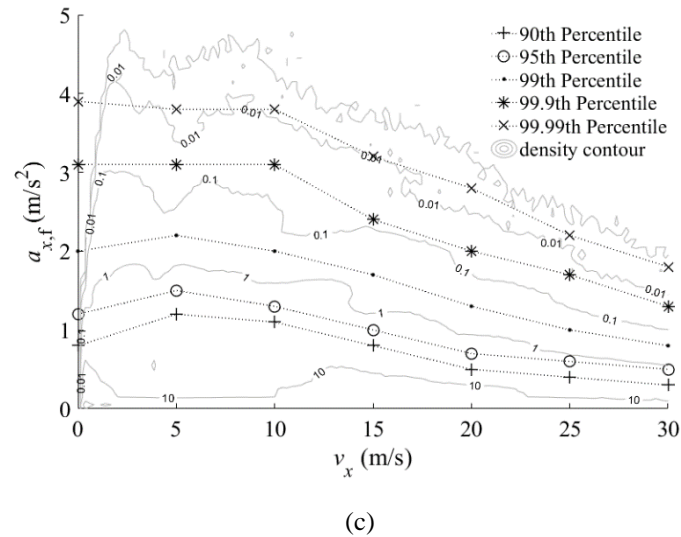

(c)

Figure 10. Percentile and relative density contour in different velocity intervals: (a) lateral acceleration, (b) brake deceleration and (c) forward acceleration.

## 4. Discussion

A Pareto distribution is usually referred to as a heavy-tail distribution, which refers to a distribution whose PDF decreases slower than that of the exponential distribution [30]. The acceleration behaviour approximating a Pareto distribution indicates that the probability of a fast acceleration is significantly higher than that of a normal distribution.

Both the longitudinal and lateral acceleration behaviours approximate a Pareto distribution. The acceleration behaviour characteristics reveal the mechanism of the quadrangle pattern, that is, the bivariate Pareto distribution and the interactions between longitudinal and lateral acceleration behaviours. Moreover, the acceleration behaviour characteristics demonstrate that the bivariate Pareto distribution model (BPDM) is an appropriate probability model for the driver. Therefore, the bivariate acceleration distribution of the driver never reaches a circle-shaped pattern. The quadrangle pattern also identifies the trade-off between the longitudinal and lateral control behaviours of the driver. The brake manoeuvre of a non-professional driver is always earlier than steering, whereas an expert driver can brake while steering after long-term training [31]. The BPDM explains why the exponential distribution is used as a priori model of the driver's acceleration behaviour in traffic simulation. The exponential distribution is a simplification of the Pareto distribution when the accuracy requirement is not high.

The interactions between the longitudinal and lateral acceleration behaviours reveal that the steering manoeuvre tends to become more intense when the brake deceleration or forward acceleration becomes larger, and the braking or accelerating manoeuvre also tends to become more intense when the lateral acceleration increases. When the drivers step on the brake or accelerator pedal strongly, they are more likely to steer the steering wheel intensely, for example, decelerating and steering at the intersection, entering or exiting the parking space, and avoiding collisions. The interactive characteristics of the acceleration behaviours explain the differences between the BPDM and NDD. The contours of the NDD are less concave because the univariate Pareto distribution of one acceleration becomes more dispersive when the other acceleration is larger. Moreover, the effect of promoting each other to move up between forward acceleration and lateral acceleration is more obvious than that between the brake deceleration and lateral acceleration. This phenomenon is also demonstrated in Figure 3. The contours in the forward acceleration section have a convex tendency, whereas the contours in the brake deceleration section have a concave tendency.

The differences between the density contours (Figure 3) and percentiles (Figure 8(a)) require attention. Figure 3 shows the 2-dimensional distribution of all data, whereas Figure 8(a) shows the percentile of lateral acceleration in a specific longitudinal acceleration interval. Figure 3 compares the absolute count numbers, whereas Figure 8(a) compares the relative frequencies. As the longitudinal acceleration increases, the absolute count number of large lateral acceleration decreases, whereas the relative frequency of large lateral acceleration increases. Therefore, both the trends shown in Figure 3 and 8(a) are reasonable.

The driver acceleration behaviours are intense at medium speed (5 m/s–10 m/s), whereas the driver acceleration behaviours are relatively gentle at high speed or low speed. At low speeds, as the velocity increases the percentiles of the accelerations increase; at high speeds, as velocity increases the percentiles of the accelerations decrease. These acceleration behaviours can be explained by the driving scenario and preview characteristics of the driver. The driver chooses a lower speed in the complex traffic scenario and a higher speed in the simple traffic scenario. There will be more emergency braking and rapid accelerating requirements in a complex traffic environment, whereas less emergency braking and rapid accelerating behaviours will be needed in a simple traffic environment. The preview characteristics of a driver can also be employed to identify this behaviour. Drivers prefer decelerating gradually to a suitable speed in advance when they preview complex driving scenarios rather than decelerating intensely as they encounter danger.

### 5. Conclusion

In this study, it is revealed that the driver's bivariate acceleration distribution presents a quadrangle pattern for two reasons: (1) the bivariate Pareto distribution and (2) the interactions between longitudinal and lateral acceleration behaviours. Therefore, the bivariate acceleration behaviour of the driver will never reach the circle-shaped region. The percentile analysis shows that the braking/accelerating manoeuvre or steering manoeuvre becomes more intense when the other acceleration is larger. This explains the difference between the BPDM and NDD results. Furthermore, the velocity has a significant influence on the acceleration behaviour of the driver. The driver acceleration behaviour is intense at medium speed (5 m/s–10 m/s), whereas it is relatively gentle at high or low speeds. These characteristics indicate that the bivariate Pareto distribution model (BPDM) is an appropriate probability model for describing bivariate acceleration behaviour. The probability model can be utilised in the human-like driving control strategy, intelligent vehicle test scenario design, and target vehicle state estimation.

**Acknowledgments**

This work was supported by the National Key Research and Development Plan of China (2018YFB1600701), Fok Yingdong Young Teachers Fund Project (171103), Key Research and Development Program of Shaanxi (2018ZDCXL-GY-05-03-01), National Natural Science Foundation of China (52002034), Innovation Capability Support Program of Shaanxi (2021TD-28), Natural Science Basic Research Program of Shaanxi (2020JQ-364), and Fundamental Research Funds for the Central Universities (300102221103, 300102229105).

**Declaration of conflicting interests**

The authors declare that there is no conflict of interest.

**Clarification for the same**

We clarify this manuscript is the same with the following preprint links:

(1)https://arxiv.org/abs/1907.01747

(2)https://www.researchgate.net/publication/334224399_Statistical_Characteristics_of_Driver_Accelerating_Behavior_and_Its_Probability_Model

(3)https://www.semanticscholar.org/paper/Statistical-Characteristics-of-Driver-Accelerating-Liu-Zhu/e08f15829039fc0a0e30911d20e9643a33b2f5d6

**Biographies:**

Rui Liu received the B.S. and M.S. degree in vehicle engineering from Chang'an University, Xi'an, China, in 2012 and 2015, respectively. He received his Ph.D. degree in vehicle engineering from Tongji University, Shanghai, China, in 2020. He is currently a lecturer with the School of Automobile, Chang'an University, Xi'an, China. His research interests include intelligent vehicle perception and decision, naturalistic driving studies, vehicle active safety, and evaluation of intelligent vehicles.

Xuan Zhao received the B.S, M.S, and Ph.D. degrees in vehicle engineering from Chang'an University, China, in 2007, 2009, and 2012, respectively, where he is currently a professor with the School of Automobile. He has undertaken over 9 government sponsored works, including the National Key Research and Development Program of China and the China Postdoctoral Science Foundation. His main research interests include intelligent vehicle perception and control, intelligent vehicle evaluation, and electric vehicle control strategy.

Xichan Zhu received the Ph.D. degree in vehicle engineering from Tsinghua University, Beijing, China, in 1995. From 1996 to 2005, he worked with China Automotive Technology and Research Center (CATARC), Tianjin, China. He was the chief expert and vice chief engineer in CATARC from 2000. He is currently working as a professor of School of Automotive Studies, Tongji University, Shanghai, China. His research interests include vehicle active safety and vehicle passive safety.

Jian Ma received the Ph.D. degree in transportation engineering from Chang'an University, China, in 2001. He is currently working as a professor with the School of Automobile, Chang'an University. He has undertaken more than 30 government-sponsored research projects, such as 863 projects and key transportation projects in China. He has published more than 90 academic articles and has authored 4 books. His main research interests include vehicle dynamics, electric vehicle and clean energy vehicle technology, and automobile detection technology and theory.